\documentclass{article}

\usepackage[ruled,vlined,algo2e]{algorithm2e}
\usepackage{algorithm}
\usepackage{algpseudocode}
\usepackage{algorithmicx}
\usepackage{algpseudocode}
\usepackage{amsmath}
\usepackage{xcolor}
\usepackage{bm}
\usepackage{arxiv}

\usepackage[utf8]{inputenc} 
\usepackage[T1]{fontenc}    
\usepackage{hyperref}       
\usepackage{url}            
\usepackage{booktabs}       
\usepackage{amsfonts}       
\usepackage{nicefrac}       
\usepackage{microtype}      
\usepackage{graphicx}
\usepackage{comment}
\usepackage{multicol}
\usepackage{multirow}
\usepackage{makecell}
\usepackage{booktabs}
\usepackage{lipsum}
\usepackage{graphicx}
\usepackage{natbib}
\usepackage{amssymb}

\usepackage{subcaption}
\usepackage{booktabs}

\graphicspath{ {./images/} }

\title{Yuan3.0 Flash: An Open Multimodal Large Language Model for Enterprise Applications}

\author{YuanLab.ai \\ research@yuanlab.ai}

\begin{document}
\maketitle
\begin{abstract}
We introduce Yuan3.0 Flash, an open-source Mixture-of-Experts (MoE) MultiModal Large Language Model featuring 3.7B activated parameters and 40B total parameters, specifically designed to enhance performance on enterprise-oriented tasks while maintaining competitive capabilities on general-purpose tasks. To address the overthinking phenomenon commonly observed in Large Reasoning Models (LRMs), we propose Reflection-aware Adaptive Policy Optimization (RAPO), a novel RL training algorithm that effectively regulates overthinking behaviors. In enterprise-oriented tasks such as retrieval-augmented generation (RAG), complex table understanding, and summarization, Yuan3.0 Flash consistently achieves superior performance. Moreover, it also demonstrates strong reasoning capabilities in domains such as mathematics, science, etc., attaining accuracy comparable to frontier model while requiring only approximately 1/4 to 1/2 of the average tokens. Yuan3.0 Flash has been fully open-sourced to facilitate further research and real-world deployment: https://github.com/Yuan-lab-LLM/Yuan3.0.

\begin{figure*}[h]
\centering
  \includegraphics[width=\linewidth]{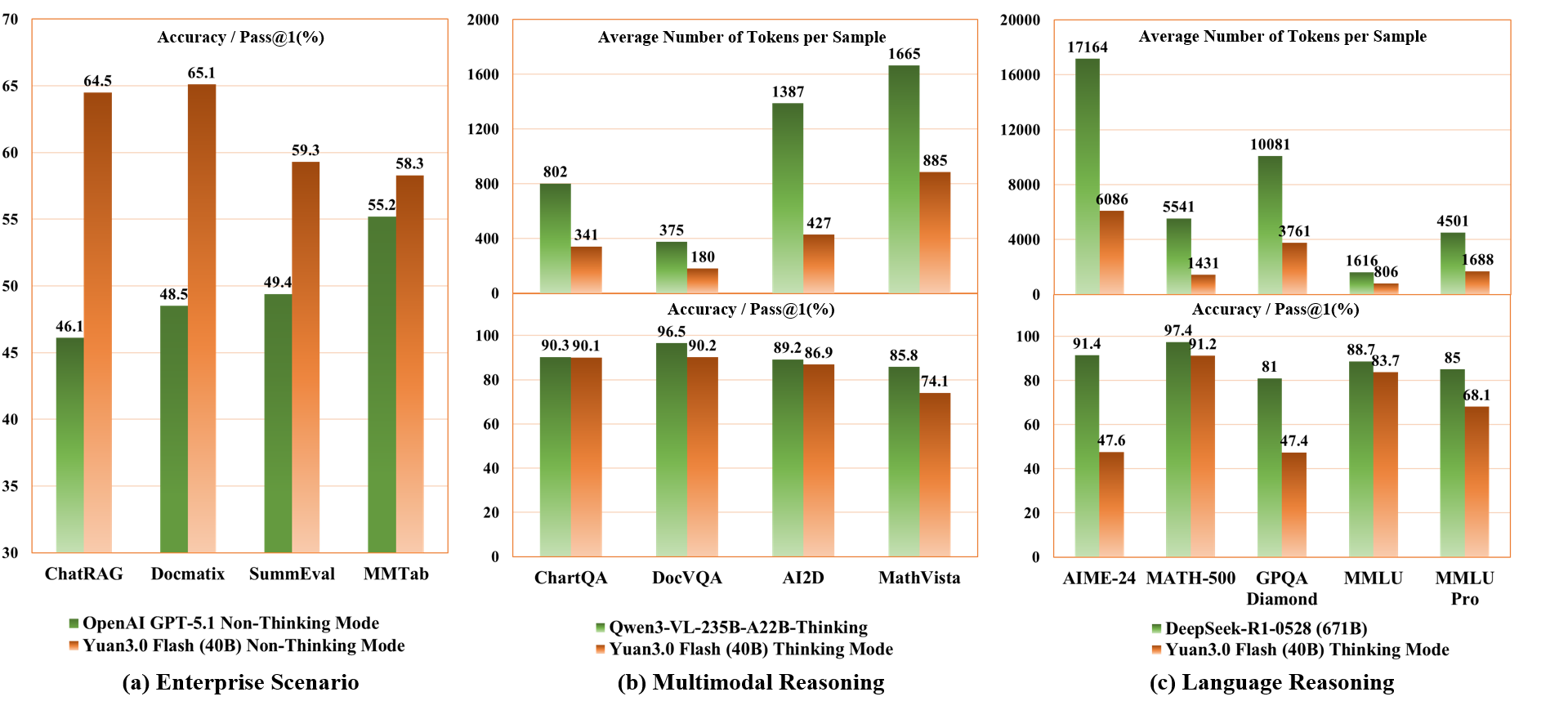}
  \caption{Performance–Efficiency Comparison of Yuan3.0 Flash across (a) Enterprise, (b) Multimodal, and (c) Language reasoning Benchmarks.}
  \label{fig:abstract}
\end{figure*}
\end{abstract}


\section{Introduction}
Large language models (LLMs) have been applied to a wide range of tasks and domains \citep{yin2024survey,liu2024deepseek}. By integrating visual and linguistic domains, Multimodal Large language models (MLLMs) unite visual perception with natural language understanding, equipped with advanced comprehension and generation capabilities to tackle complex multimodal tasks   \citep{wu2023multimodal,hurst2024gpt,bai2025qwen2,comanici2025gemini,chen2024internvl}. Despite the rapid development of LLMs and MLLMs \cite{comanici2025gemini,guo2025deepseek}, there are still shortcomings in key capabilities such as RAG, multimodal document processing, complex table analysis, summary generation, etc., when applying them to enterprise scenarios. These deficiencies directly limit their application in enterprise-level scenarios such as Intelligent Customer Service, R\&D digitization, marketing analysis, etc.

Long chain-of-thought (CoT) reasoning has become a dominant research paradigm for both unimodal LLMs and MLLMs, driving significant advances in complex problem-solving \citep{chen2025towards,guo2025deepseek,jaech2024openai}. The rapid evolution of MLLMs has extended the CoT paradigm to multimodal tasks, where models are required to reason over and integrate information from heterogeneous inputs like images, charts, and text  \citep{liu2024improved,zhang2023multimodal}. These models, by generating rationales that bridge visual perception and linguistic inference, have demonstrated remarkable capabilities in visual question answering, document analysis, and complex reasoning. However, this long CoT paradigm introduces the overthinking problem, where models generate excessively long reasoning traces, contributing to the waste of computational resources \citep{lou2025adacot,liu2025think}.

In order to improve key capabilities in enterprise applications, and alleviate overthinking issue in LRMs, we carry out the following work:

\begin{itemize}
    \item Develop a high-quality data generation system, which enables filtering of diverse multimodal data types, and generates high-quality pre-training and post-training datasets for vertical domains such as manufacture, finance, law, education, healthcare, etc.
    \item Propose a reflection-aware policy optimization (RAPO) reinforcement learning algorithm that effectively mitigates the overthinking issue of  LRMs, and improves training efficiency and model accuracy.
    \item Train a multimodal model Yuan3.0 Flash with 40B parameters, which achieves superior performance on a wide range of enterprise-oriented benchmarks, and demonstrates strong capabilities in general tasks.
       
\end{itemize}

\section{Model Architecture}
\begin{figure*}[t]
  \includegraphics[width=\linewidth]{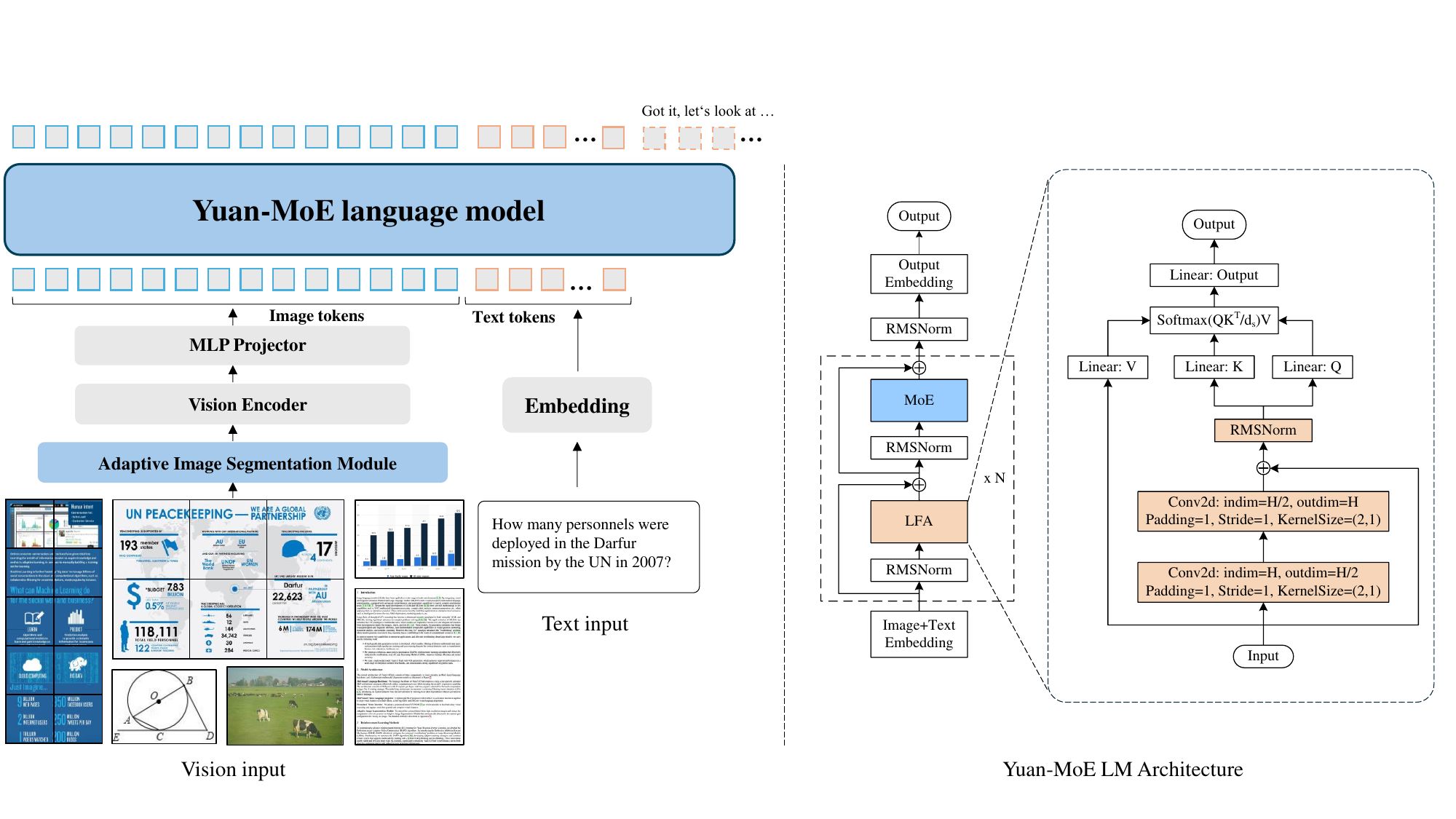}
  \caption{Overall architecture of Yuan3.0 Flash and MoE-based Language Backbone. The left figure depicts the proposed Yuan3.0 architecture, which comprises three integral components: (1) a ViT Encoder responsible for processing and encoding input images; (2) a lightweight MLP projector with SwiGLU activations to align visual features with the textual token space; and (3) a MoE-based Language Model that serves as the Language Decoder. The right figure presents the language backbone with Localizing Filtering-based Attention (LFA)}
  \label{fig:structure}
\end{figure*}

The overall architecture of Yuan3.0 Flash consists of three components: a visual encoder, an MoE-based language backbone, and a lightweight multimodal alignment module as illustrated in Figure \ref{fig:structure}.

\textbf{MoE-based Language Backbone}: The language backbone of Yuan3.0 Flash employs a large-scale sparsely activated MoE architecture, aiming to effectively reduce computational costs while ensuring the model's expressive capability. The architecture consists of 40 layers with 32 experts per layer, with two experts selected for forward computation using a Top-K routing strategy. The underlying architecture incorporates Localizing Filtering-based Attention (LFA) \citep{wu2023yuan}, introducing an explicit inductive bias into self-attention by favoring local token dependencies that are prevalent in natural language.

\textbf{MLP-based Vision-Language projector}: A lightweight MLP projector with SwiGLU as activation function is applied to align visual features to textual tokens, achieving stable and efficient visual-language alignment.

\textbf{Pretrained Vision Encoder}: We adopt a pretrained InternViT-300M \citep{chen2024internvl} as vision encoder to facilitate deep visual reasoning and capture more fine-grained and complex visual features.

\textbf{Adaptive Image Segmentation Module}: To extract fine-grained detail from high-resolution images and reduce the computation costs, we propose an Adaptive Image Segmentation Module that automatically determines the optimal grid configuration for slicing an image. The detailed method is described in Appendix \ref{app:seg}.

\section{Reinforcement Learning Methods}

To address the overthinking issue and systematically advance reinforcement learning (RL) training for Yuan3.0 across diverse scenarios, we develop the Reflection-aware Adaptive Policy Optimization (RAPO) algorithm. By introducing the Reflection Inhibition Reward Mechanism (RIRM), RAPO effectively mitigates the common "overthinking" problem in LRMs. Furthermore, we improve the DAPO algorithm \citep{yu2025dapo} by introducing adaptive training strategies and a unified reward system that supports multi-task RL training with a hybrid of thinking and non-thinking mode. These innovations significantly enhance Yuan3.0 Flash's performance with stable and efficient large-scale RL training.

\subsection{Reflection Inhibition Reward Mechanism (RIRM)}
LRMs driven by reinforcement learning with verifiable rewards (RLVR) generate long reasoning chains to enhance inference capabilities. Models such as OpenAI's o1 \citep{jaech2024openai}, DeepSeek-R1 \citep{guo2025deepseek}, and Kimi-1.5 \citep{team2025kimi} have demonstrated the effectiveness of this approach. However, the RLVR mechanism tends to induce overthinking, which not only wastes computational resources but can also degrade model performance. This issue has become a critical bottleneck in practical LRM applications.

\begin{figure*}[t]
  \includegraphics[width=\linewidth]{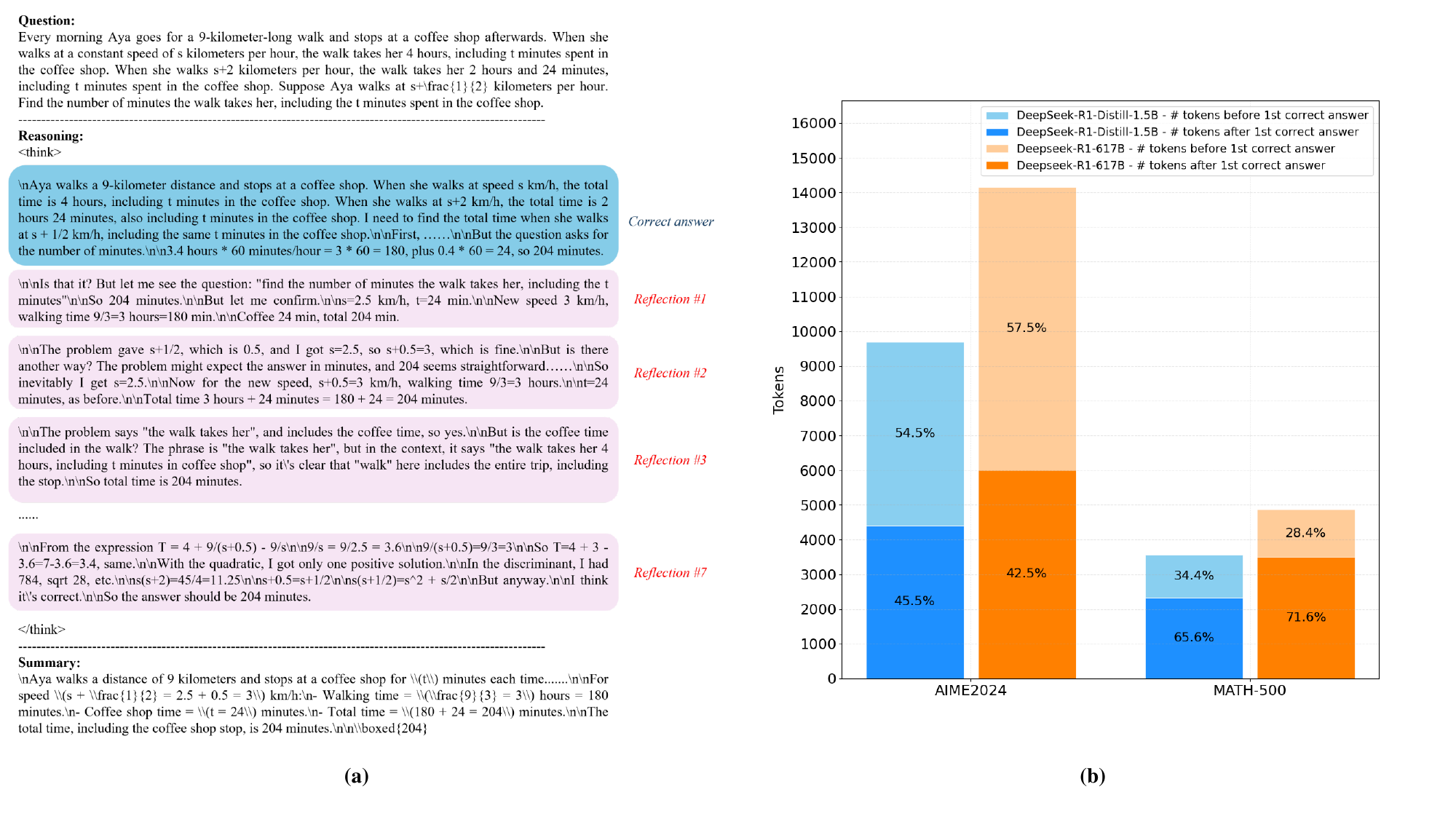}
  \caption{(a) An example of Deepseek-R1's reasoning process, showing repetitive "reflection" behavior after deriving the "first answer". (b) Average token consumption breakdown of DeepseekR1-Distill-1.5B and Deepseek-R1 across AIME 2024 and MATH-500 benchmarks. The light-colored segments indicate token consumption before the "first answer" appears, while the dark-colored segments correspond to token consumption during the "reflection" phase. }
  \label{fig:ds_distill}
\end{figure*}

In complex reasoning tasks—particularly in mathematical and scientific domains—LRMs exhibit pronounced overthinking behavior. As illustrated in Figure \ref{fig:ds_distill}(a), after correctly deriving an answer through a complete logical chain, models engage in repetitive self-checking and verification—a phenomenon  termed "reflection". Our statistical analysis of Deepseek-R1 and DeepseekR1-Distill-1.5B \citep{guo2025deepseek} model on the AIME 2024 and MATH-500 benchmarks reveals that up to 71.6\% of token consumption occurs in the post-answer reflection phase. (Figure \ref{fig:ds_distill}(b)).

To mitigate overthinking problem, we propose the RIRM. RIRM identifies the step where the model initially arrives at the correct answer and its subsequent reflection steps during the reasoning process. It then assesses the reasoning trajectory along three key dimensions: whether the first correct answer is identified, whether the number of reflection steps is reasonable, and whether the final answer is correct. These dimensions are aggregated into a reward signal to guide policy optimization in reinforcement learning. 

\begin{figure}[h]
  \includegraphics[width=\linewidth]{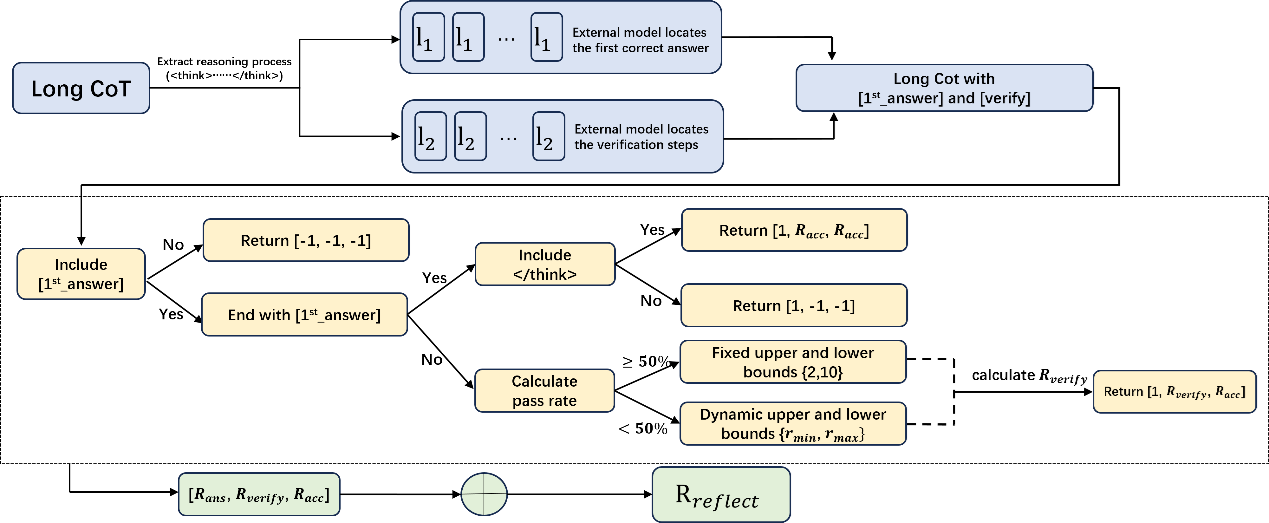}
  \caption{Illustration of Reflection Inhibition Reward Mechanism. }
  \label{fig:RIRM}
\end{figure}

As shown in Figure \ref{fig:RIRM}, RIRM is implemented in three stages:

1. Extract reasoning trajectory from the model's output, split them into equal-length strings, and use an LLM to annotate where the correct answer first appears and reflection steps with “[$1^{st}$\_answer]” and “[verify]” respectively.
    
2. Calculate three scores using the annotated reasoning trajectory:
\begin{itemize}
    \item $R_{ans}$: whether the first correct answer is identified.
    \item $R_{ver}$: whether the number of reflections, $v$, is reasonable. For $R_{ver}$ calculation, samples with $\ge 50\%$ accuracy rate use fixed bounds $r_{min}=2$, $r_{max}$=10, while others use dynamic bounds $r_{min}=\tilde{r} _{min}$, $r_{max}=\tilde{r} _{max}$.  $\tilde{r} _{min}$ and $\tilde{r} _{max}$ are the minimum and maximum number of “[verify]” among N samples of the same prompt. $R_{ver}$ is calculated as:
        \begin{align}
            R_{ver}(v) = 
            \begin{cases}
            1& v\leq r_{min},\\
            1-\frac{(v-r_{min})}{(r_{max}-r_{min})} & r_{min}<v\leq r_{max},\\
            0& r_{max} < v.
            \end{cases}
        \end{align}
    \item $R_{acc}$: whether the final answer is correct.
\end{itemize}

3. The three component scores:[$R_{ans}$, $R_{ver}$, $R_{acc}$ ] are summed to obtain the final reflection inhibition reward:
\begin{equation}
    R_{reflect}=R_{ans}+R_{ver}+R_{acc}
\end{equation}    

\begin{figure}[h]
  \includegraphics[width=\linewidth]{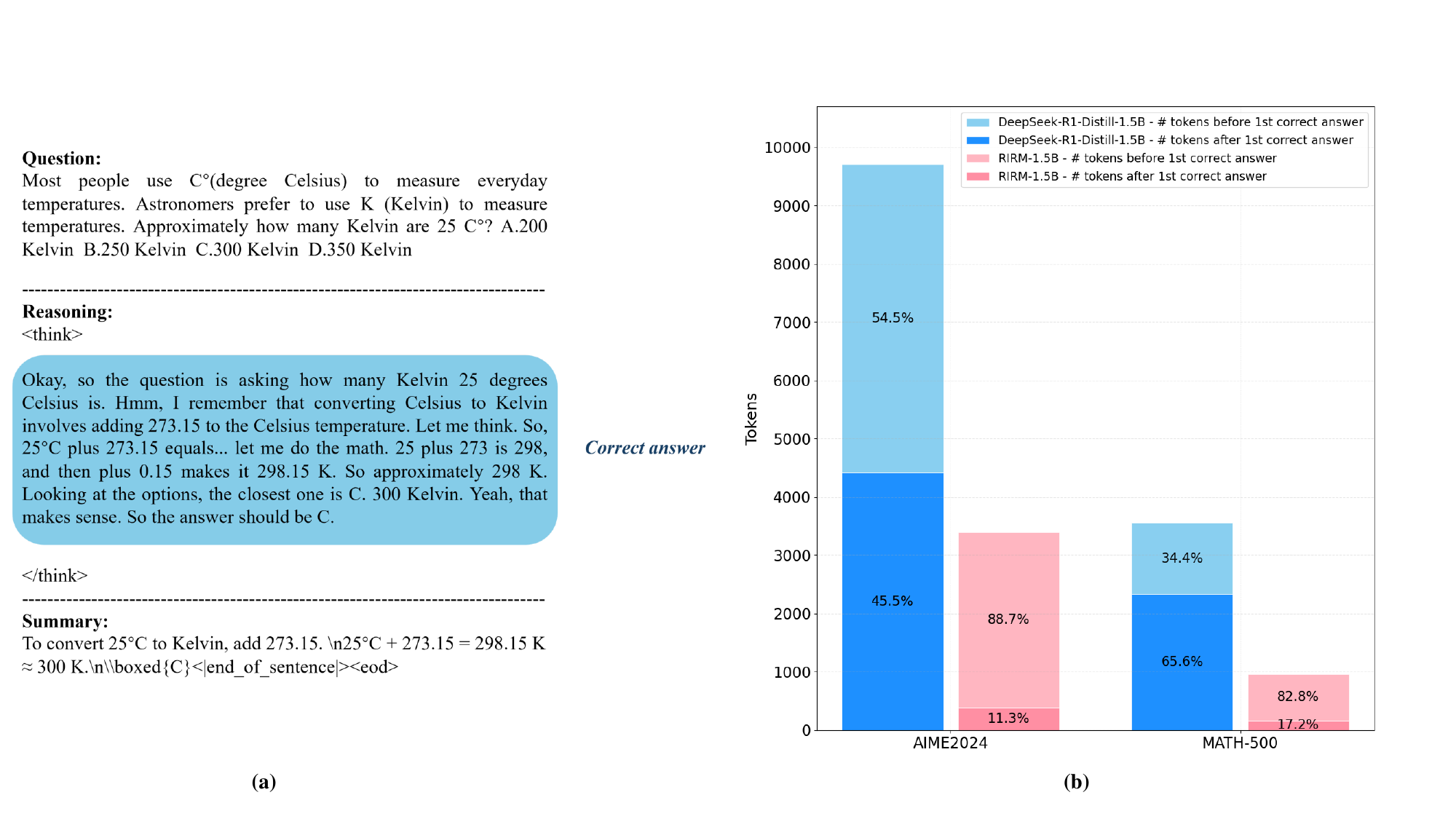}
  \caption{(a) Example of DeepseekR1-Distill-1.5B’s reasoning process after trained with RIRM. (b) Comparison of average token consumption of DeepseekR1-Distill-1.5B before and after trained with RIRM in AIME 2024 and MATH-500 benchmarks. The light-colored segments indicate token consumption before the "first answer" appears, while the dark-colored segments correspond to token consumption during the "reflection" phase. }
  \label{fig:ds_reasoning}
\end{figure}

The incorporation of  RIRM effectively mitigates overthinking issue and improve model accuracy for Yuan3.0 Flash. Table \ref{tab:RLcomparison} presents a comparison of accuracy and output tokens on AIME 2024 and MATH-500 between the Yuan3.0 Flash 40B SFT base model trained with DAPO length penalty and with RIRM. Compared to the baseline model, the RIRM-based reinforcement learning framework yields a maximum accuracy improvement of 52.37\% while simultaneously reducing output tokens by up to 47.14\%. Figure \ref{fig:ds_reasoning} demonstrates the effects of RL training with RIRM, showing that the reasoning process of DeepseekR1-Distill-1.5B becomes significantly more concise. Notably, the overall token consumption is reduced by up to 64.09\%, and the token consumption during the reflection phase see a reduction of up to 90.58\%.

\begin{table}[h]
\centering
\caption{Performance comparison of the Yuan3.0 Flash (40B) model under RL training with explicit DAPO length penalty versus the proposed RIRM.}
\begin{tabular}{lcccc}
\toprule 
\multirow{2}{*}{Method} & \multicolumn{2}{c}{AIME 2024} & \multicolumn{2}{c}{MATH-500} \\
\cline{2-5}
                        & Accuracy & Tokens & Accuracy & Tokens \\
\midrule
Yuan3.0 Flash (40B) SFT        & 31.45    & 13,656  & 83.20    & 3,362   \\
RL+DAPO length-penalty  & 46.35    & 13,781  & 89.06    & 3,974   \\
RL+RIRM                 & \textbf{47.92}    & \textbf{7,505}   & \textbf{89.47}    & \textbf{1,777}   \\
\bottomrule
\end{tabular}
  \label{tab:RLcomparison}
\end{table}

\subsection{Optimized DAPO}
DAPO removes the KL penalty term and integrates key techniques—including the clip-higher mechanism, dynamic sampling, token-level policy gradient loss, and reward shaping for ultra-long sequences—to form a superior optimization objective:
\begin{align}
\mathcal{J}_{\text{DAPO}}(\theta) =& \mathbb{E}_{(q,a)\sim\mathcal{D},\{o_i\}_{i=1}^G\sim\pi_{\theta_{old}}(\cdot|q)}\nonumber\\
&\Bigg[\frac{1}{\sum_{i=1}^G|o_i|}\sum_{i=1}^G\sum_{t=1}^{|o_i|}\min\Big(r_{i,t}(\theta)\hat{A}_{i,t},\text{clip}\big(r_{i,t}(\theta),1-\epsilon_{\text{low}},1+\epsilon_{\text{high}}\big)\hat{A}_{i,t}\Big)\Bigg],\nonumber\\
&\text{s.t. } 0<\big|\{o_i|\text{is\_equivalent}(a,o_i)\}\big|<G
\end{align}

To further enhance computational performance and stabilize gradient updates, we have introduced a series of optimizations to the DAPO algorithm.

\textbf{Adaptive Dynamic Sampling}\quad   DAPO's dynamic sampling method oversamples and filters out prompts with identical reward scores, to ensure all prompts in a batch have valid gradients while maintaining a consistent prompt count. However, it typically requires sampling three times the training batch size to guarantee sufficient valid samples, with excess samples being discarded. This leads to data waste, increases training iteration time, and increase the need for additional training epochs.

\begin{figure*}[t]
  \includegraphics[width=0.48\linewidth]{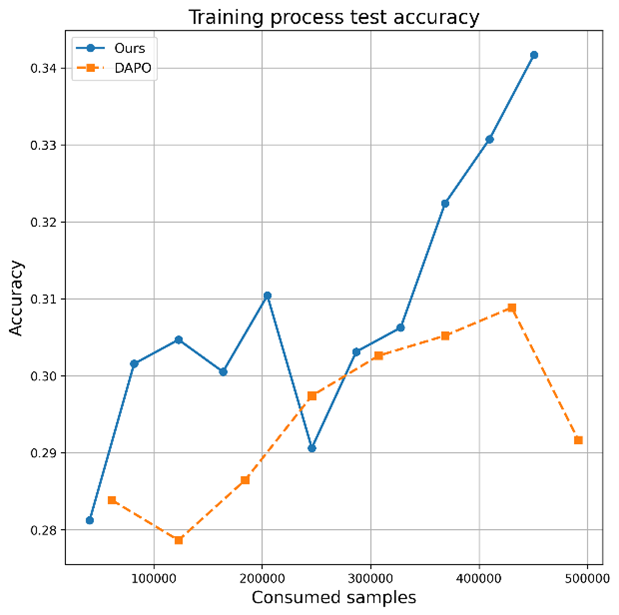} \hfill
  \includegraphics[width=0.48\linewidth]{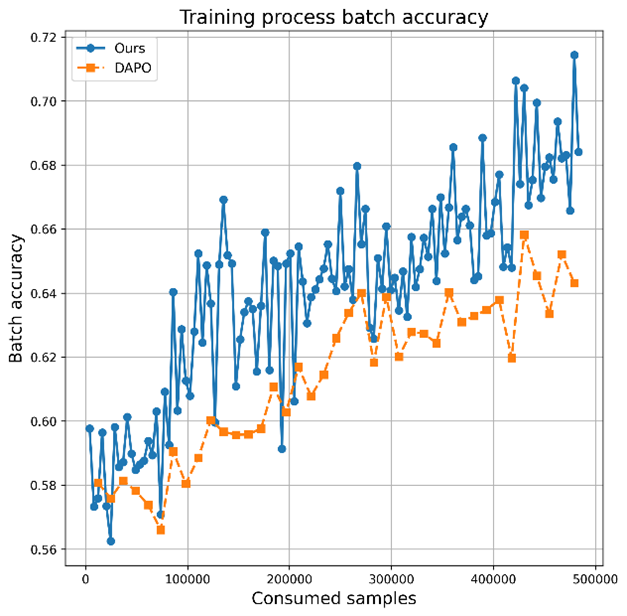}
  \caption {Training and testing accuracy of DeepSeek-R1-Distill-Qwen-1.5B under DAPO with and without ADS.}
  \label{fig:dapoads}
\end{figure*}

To improve sampling efficiency, we develop the Adaptive Dynamic Sampling (ADS) strategy that refines DAPO's dynamic sampling process. We first sample a batch matching the training batch size and filter out prompts with identical rewards. For the remaining prompts, we compute their passing rates: $P_j= \frac{1}{G} { \sum_{i=1}^{G}} I (a_i =1)$. We then sort prompts by passing rate in descending order and resample from the highest-pass-rate prompts to replenish the batch. To control resampling intensity, we replenish the batch to the smallest multiple of the mini-batch size, ensuring sufficient samples for training without excessive duplication. For detailed algorithm specifications, please refer to the Appendix \ref{app:ADS}. As shown in Table \ref{tab:dapoads}, DAPO with ADS achieves a significant reduction in both average sampling time and per‑step iteration time compared to the baseline DAPO, resulting in a 52.91\% improvement in training efficiency. Correspondingly, Figure \ref{fig:dapoads} illustrates the model’s consistent accuracy improvement throughout both training and testing phases.

\begin{table}
  \centering
   \caption{DAPO enhanced with ADS substantially accelerates training of DeepSeek-R1-Distill-Qwen-1.5B compared to the baseline.}
  \begin{tabular}{lcc}
    \toprule
    Case  &\makecell[c]{Average time \\ of generation (s)}   &\makecell[c]{Average time \\ of step (s)}\\
    \midrule
    DAPO     & 455.73      &928.48     \\
    DAPO+ADS    & \textbf{257.22}     &\textbf{437.18}      \\
   \bottomrule
  \end{tabular}
  \label{tab:dapoads}
\end{table}

\textbf{80/20 rule}\quad   DAPO encounters training instability issues when applied to large-scale Mixture-of-Experts (MoE) models. Building upon DAPO, we apply the 80/20 algorithm \citep{wang2025beyond}, which updates gradients using only the top 20\% of tokens with the highest entropy during RL training. This approach not only significantly enhances training stability but also improves model accuracy.

\textbf{Optimized dual-clip}\quad   During large-scale RL training, the updated policy model may significantly deviate from the old policy, resulting in abnormally large probability ratios between new and old policies for certain trajectory data. When the advantage values are negative, these extreme ratios can lead to unbounded gradients, causing model collapse. While dual-clip PPO \citep{ye2020mastering} is commonly used to mitigate such instability, it introduces an additional hyperparameter $c$. Improper tuning of $c$ may hinder model convergence, and the extra bounds checking on the product of probability ratios and advantage values increases computational overhead as training data scales up. To address this, we optimize the standard PPO's clipped probability ratio approach: when encountering abnormally large probability ratios with negative advantages, we exclusively use the clipped ratio to prevent extreme gradient values, thereby further stabilizing gradient updates.

Integrating the aforementioned optimizations, we compute the RL loss using the following formula: 
\begin{align}
    J^{\color{red}\mathcal{B}_r}=&\mathbb{E}_{\color{red}\mathcal{B}_r\sim\mathcal{D},(q,a)\sim\mathcal{B}_r,\color{black}\{o_i\}_{i=1}^G\sim\pi_{\theta_{old}}(\cdot|q)}\nonumber\\
    &\Bigg[\frac{1}{\sum_{i=1}^G|o_i|}\sum_{i=1}^G\sum_{t=1}^{|o_i|}\color{red}\mathbb{I}\Big[H_{t}^i\ge\tau_{\rho}\Big]\color{red}PG_{i,t,\epsilon_{low},\epsilon_{high}}(\theta)\color{black}\Bigg],\text{s.t. } \color{red}|\mathcal{B}_r| = k\cdot |\mathcal{B}_{\text{mini}}|
\end{align}

where
\begin{align}
    PG_{i,t,\epsilon_{low},\epsilon_{high}}(\theta)=
    \begin{cases}
        \text{clip}\big(r_{i,t}(\theta),1-\epsilon_{\text{low}},1+\epsilon_{\text{high}}\big)\hat{A}_{i,t},
            &\text{ if }\hat{A}_{i,t}\le0,\\
        \begin{aligned}[t]
            \min\Big(r_{i,t}(\theta)\hat{A}_{i,t},\text{clip}\big(r_{i,t}(\theta),1-\epsilon_{\text{low}},1+\epsilon_{\text{high}}\big)\hat{A}_{i,t}\Big)
        \end{aligned}
            & \quad\quad o.w.
    \end{cases}
\end{align}

In the equations, red text highlights our algorithmic optimizations. $H_i^t$ denotes the entropy of token $t$ in response $i$. $\mathbb{I}[\cdot]$ is the indicator function, which takes the value 1 if the condition holds and 0 otherwise. $\rho=0.2$ is the top proportion of high‑entropy tokens we specify for each response. $\tau_{\rho}$ is the corresponding entropy threshold, such that only tokens with $H_i^t \ge \tau_{\rho}$ are used for gradient computation; these tokens constitute the top $\rho$ portion of all tokens in the response. $\mathcal{B}_{mini}$ denotes the mini‑batch size, and after adaptive dynamic sampling, the global batch size $\mathcal{B}_{r}$ becomes $k$ times the mini‑batch size.

\begin{figure*}[t]
  \includegraphics[width=\linewidth]{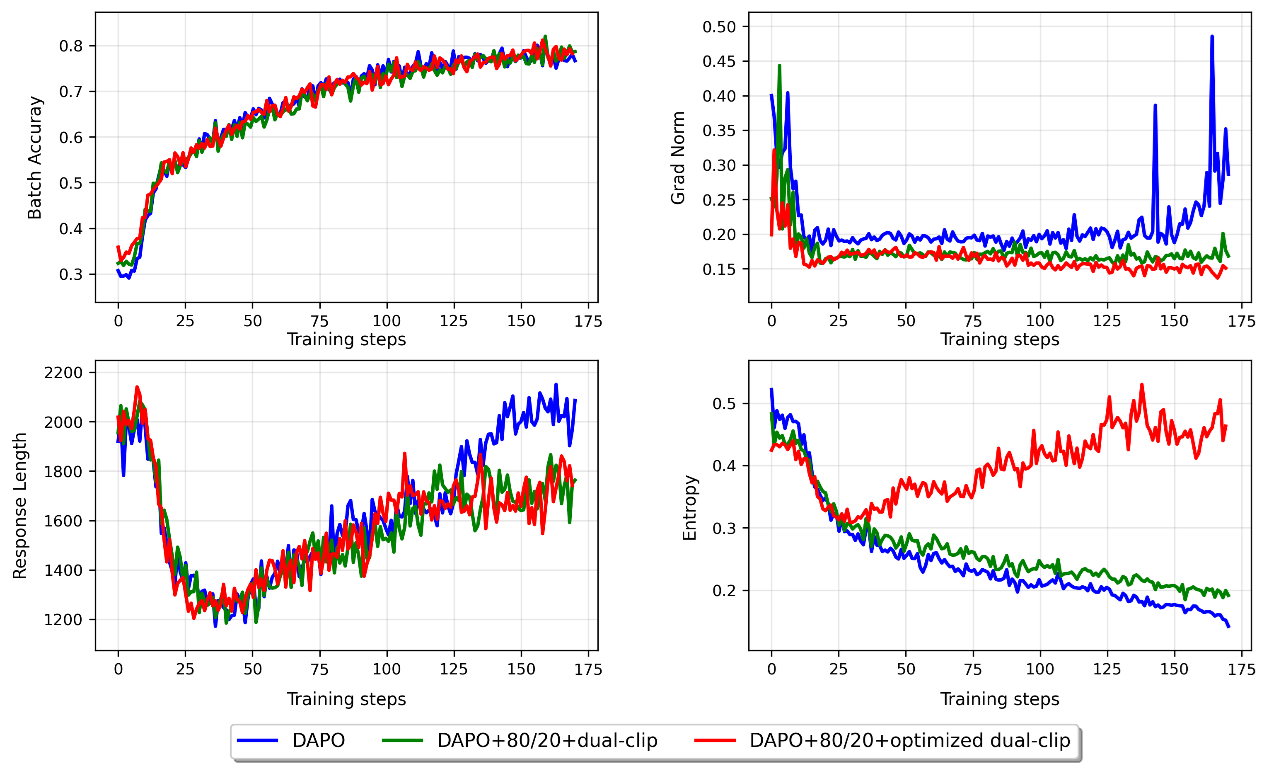}
  \caption{Training dynamics of DAPO (blue line), DAPO with 80/20 rule and dual-clip technique (green line), and DAPO with 80/20 rule and our optimized dual-clip technique. }
  \label{fig:train_dynamic}
\end{figure*}

The optimized DAPO algorithm (Figure \ref{fig:train_dynamic}) maintains the lowest gradient norm, indicating high training stability. This contrasts with DAPO without the 80/20 rule, whose gradient norm oscillates and sharply increases in later stages, leading to rapid collapse. The stability of our approach is also evidenced by moderate length growth and a gradual entropy increase, a pattern correlated with better model performance as suggested in \cite{yu2025dapo}.

\subsection{Multi-Task Training with Adaptive Strategies}
Unlike Deepseek-R1 or Qwen 3 series models, which strictly separate deep thinking training from fast thinking training, or Kimi1.5 which trains two separate models for these distinct capabilities, we adopt a \textbf{mixed training paradigm} that jointly trains a single model on both thinking tasks (including language and multimodal mathematical reasoning, code generation, and scientific tasks) and non-thinking tasks (such as language and multimodal RAG, language and multimodal chat, and general scenario QA). This unified approach thus enables the model to achieve superior generalization and adaptability across a wide range of domains.

To preserve task-specific characteristics, we employ distinct training strategies:
\begin{itemize}
    \item For deep thinking tasks involving language/multimodal mathematical reasoning and scientific tasks, we utilize RIRM.
    \item For code generation under thinking tasks and all non-thinking tasks, we apply optimized DAPO with explicit length constraints and penalties
\end{itemize}

Different maximum output limits and penalty intervals are carefully configured for each task category to prevent interference and ensure training stability.

\textbf{Data grouping and alternating training strategy}\quad    When data from different tasks have widely varying maximum output lengths (e.g., 4K vs. 16K), traditional mixed batching would force shorter sequences to idle while waiting for the longest one in a batch, severely hampering efficiency. To address this, we introduce a data grouping and alternating training strategy based on maximum output lengths: data are first grouped by length categories (e.g., 4K group, 16K group), and then trained alternately according to their proportion (e.g., with a 4K:16K ratio of 2:1, two 4K batches are followed by one 16K batch). Experiments on the Yuan3.0 Flash 40B model show that this method improves inference phase throughput by 16\%.

\textbf{Repetition truncation}\quad    Furthermore, to maintain output quality during reinforcement learning, we implement repetition truncation in trajectory collection. The model’s output is monitored in real time, and every 1024 tokens generated undergo repetition detection. If any continuous subsequence of length N repeats beyond a preset threshold, the episode is terminated early to prevent degenerate output loops. In our setup, N and the repetition threshold are set to 200 and 10, respectively.

\subsection{Unified Reward System: Verifiable Framework and Generative Reward Models}
To comprehensively guide model optimization across both objective and open-ended tasks, we establish a Unified Reward System consisting of two complementary components: the Verifiable Reward Framework for tasks with deterministic criteria, and Generative Reward Models (GRMs) for open-ended scenarios requiring nuanced human-like judgment.

\textbf{Verifiable Reward Framework}\quad   The framework provides precise, rule-based evaluation across specialized domains:

\textit{Mathematics \& Science}. For mathematical reasoning and scientific tasks (covering set operations, chart interpretation, and multiple-choice questions), we employ a hybrid evaluation approach: rule-based string matching for preliminary scoring, augmented by the "math\_verify" technique to rigorously validate the equivalence of mathematical expressions and formula-based answers.

\textit{Programming Languages}:
\begin{itemize}
    \item Python: Our automated testing framework executes unit tests under multiple scenarios: standard input simulation, direct code execution, and text-embedded code extraction. Leveraging Uvicorn for asynchronous execution enables high-throughput evaluation.
    \item Generated statements are executed against database engines, with correctness determined by result consistency with reference queries.
\end{itemize}

\textit{Retrieval-Augmented Generation (RAG)}.The RAG evaluation system employs a multi-dimensional strategy combining:
\begin{itemize}
    \item Objective metrics: Threshold judgments, linear mapping, and exact matching for factual consistency
    \item Semantic evaluation: Instruction adherence and semantic understanding via LLM discriminators for quality ranking and classification
    \item Text similarity: Normalized mapping of similarity metrics for generation quality assessment
\end{itemize}

\textit{Output Quality Constraints}. To enforce standardization, we implement a multi-faceted negative reward mechanism:
\begin{itemize}
    \item Formatting Control: Regex-based validation of thought-process tags ("<think>...</think>") and structural markers ("<|Assistant|>", "<|end\_of\_sentence|>"), with -1.0 penalty for deviations
    \item Language Consistency: Character-level analysis of Chinese/English ratios (post number/symbol filtering), applying -1.0 penalty when non-target language exceeds 20\%
    \item Repetition Penalty: N-gram sliding window detection of token repetition patterns, triggering -1.0 penalty when unique character falls below 80\%.
\end{itemize}

\textbf{Generative Reward Models (GRMs)}\quad To address the lack of definitive answers in open-ended and subjective tasks, we employ Generative Reward Models (GRMs) for task-specific scoring. Compared to scalar-based and semi-scalar approaches, GRMs offer superior interpretability and generalization by translating advanced language understanding and visual perception into quantifiable rewards \citep{liu2025inference}. We train both pure-language and multimodal GRMs using Formatted Supervised Fine-Tuning (SFT) and Rejection Sampling Fine-Tuning (RFT), transforming the model's comprehension into rigorous judging capabilities through instance-specific scoring guidelines and iterative refinement.

\textit{Pure-Language GRM}.    We curate approximately 1M seed instruction-tuning samples spanning multi-turn dialogue, safety, open-ended Q\&A, riddles, text processing, and social sciences (politics, history, religion, literature, art, economics). For each query, we generate 7 diverse responses using Yuan3.0 Flash 40B, mixing them with SFT answers to form candidate lists. A teacher model evaluates these responses based on predefined standards, reasoning, and a 1–10 score.
To ensure reward accuracy, we apply ground-truth consistency filtering—only retaining samples where the original answer uniquely achieves the highest score. The rejection sampling phase further enhances precision by incorporating error-marked data, outputs from larger models, and open-source preference datasets (UltraFeedback \citep{cui2023ultrafeedback}, OffsetBias \citep{park2024offsetbias}, Skywork-Reward \citep{liu2024skywork}, HelpSteer2 \citep{wang2024helpsteer2}, JudgeLM \citep{zhu2023judgelm}. The fine-tuned reward model performs three inferences per sample, with ground truth filtering reapplied to retain high-quality data for final training.

\textit{Multimodal GRM (MGRM)}. MGRM extends evaluation to cross-modal semantic alignment, focusing on image content, chart logic, and on-screen information. Optimized for tasks including image captioning, chart analysis, OCR, code reasoning, and screenshot analysis, MGRM trains on a combination of internal data and open-source preference datasets (RLAIF-V \citep{yu2025rlaif}, RLHF-V \citep{yu2024rlhf}, POVID \citep{zhou2024aligning}, MM-RLHF \citep{zhang2025mm}, VL-Feedback \citep{li2024vlfeedback}, wildvision-battle \citep{lu2024wildvision}. A multimodal teacher model generates formatted scoring data to enhance the model's assessment of semantic accuracy, logical correctness, and descriptive clarity.
RFT further improves MGRM's adaptation to larger model distributions and data diversity. The fine-tuned model performs secondary inference and filtering on error data, with targeted debiasing (hinted sampling) ensuring stable, human-aligned feedback within the RL loop.

\section{Datasets}
\subsection{Pre-training Data}
A mixed corpus of text and image-text pairs was built to train the Yuan3.0 Base model.
\subsubsection{Textual Datasets}
\begin{figure}[t]
  \centering
  \includegraphics[width=0.7\linewidth]{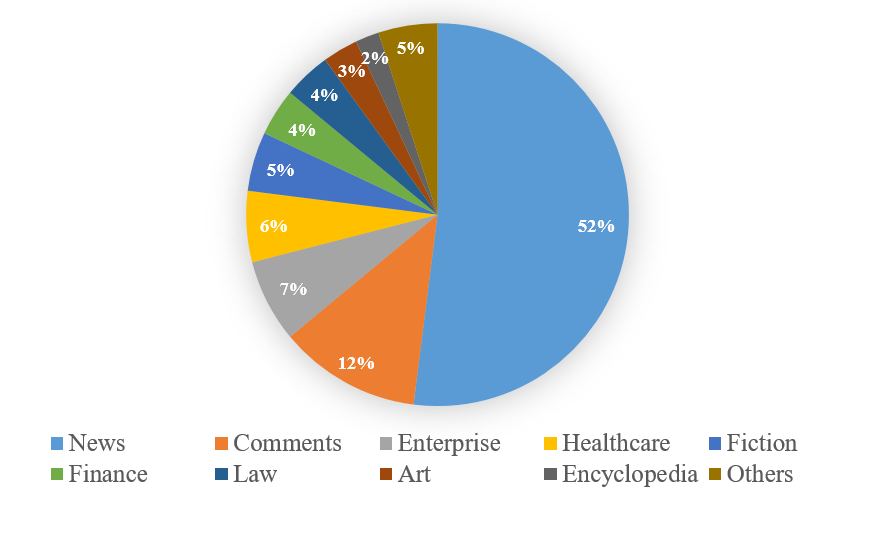}
  \caption{Domain-wise composition of the web-crawled pre-training corpus.}
  \label{fig:webdata}
\end{figure}

The pre-training textual data with 3.5TB tokens is sourced from diverse origins - including web pages, encyclopedia entries, book excerpts, academic papers and code.
To guide the curation of web-crawled data, we first traine a FastText-based domain classifier using a labeled domain dataset. This classifier is capable of distinguishing across a broad spectrum of domains, including advertising, economy, healthcare, law, business, etc. Through sampling analysis of open-source web-crawled datasets \citep{li2024datacomp} by our domain classifier, it is indicated that advertisements, lifestyle news, and entertainment/sports news accounts for the majority (Figure \ref{fig:webdata}).

We reduce the proportion of low-value news and advertising content while keeping high-value data, such as technology, science, knowledge, etc. Furthermore, we increase the proportion of data from vertical domains, such as finance, law, manufacturing, healthcare, etc. We also train a quality-scoring classifier and remove all text samples with a normalized quality score below 0.6 (on a 0–1 scale) to ensure data quality.

\subsubsection{Multi-modal Datasets}
The multimodal pre-training of Yuan3.0 Base is designed to establish robust visual-linguistic alignment and reasoning capabilities. We construct a large-scale, diverse dataset of 1.5B image-text pairs. This comprehensive corpus enables the model to learn deep semantic relationships between visual content and textual descriptions, forming a foundation for sophisticated, human-like multimodal reasoning.

\subsection{Fine-tuning datasets}

\subsubsection{ Datasets for General tasks}
For general dialogue tasks such as casual chat as well as domain knowledge QA, a dedicated scoring model is trained to construct a multi-dimensional evaluation system focusing on the sample quality. The model conducts comprehensive evaluation focusing on four key dimensions: verifying whether the answer accurately matches the user's question without deviating from the topic or being irrelevant, assessing the coherence and organization of the answer to ensure clear reasoning and logical closure, ensuring responses comply with the inherent constraints strictly removing vulgar, non-compliant, and misleading content. Only samples with scores passing the preset threshold are retained. The multi-modal dataset spanning multiple domains, including documents, charts, visual reasoning problems, code, OCR, science problems, etc., is constructed. 

\subsubsection{Datasets for Enterprise Scenario}
We construct a hybrid dataset highly relevant to enterprise application scenarios, including RAG, tabular, etc., by integrating human annotations, synthetic generation, and open-source resources. We create a 45K-dialogues dataset (25K text-only, 20K multimodal) by human expert to keep multimodal samples aligned with visual inputs including code snippets, diagrams, API screenshots, etc. We generate synthetic data across wide range of domains including Literature, Geography, Biology, Economics, etc., leveraging data sources such as Wikipedia, arXiv, etc. The synthetic data includes 305K samples (160K text-only, 145K multimodal), which ensures semantic diversity, covers 95\% of low-frequency concepts. We also build general instruction datasets \citep{wei2021finetuned,taori2023stanford} and multimodal understanding datasets \citep{goyal2017making,hudson2019gqa,singh2019towards,mishra2019ocr,lin2014microsoft,krishna2017visual,liu2023visual} covering 200K text samples and 160K multimodal samples.

\subsection{Datasets for Reinforcement Learning}

We have designed a standardized workflow to generate reinforcement learning datasets for a wide range of domains. Firstly, data with ground truth answers are extracted as candidate data. For code data, the answer comprises several unit test cases (with more than 5 unit tests). Secondly, for mathematical and scientific data, a rule-based filtering is applied to remove samples including multiple choice and proof-based questions. Next, the remaining samples are graded by the pass rate of Yuan3.0 Flash SFT model. Only those difficult samples are retained for subsequent reinforcement learning training. 

\section{Training Pipeline}

\begin{figure}[t]
  \includegraphics[width=\linewidth]{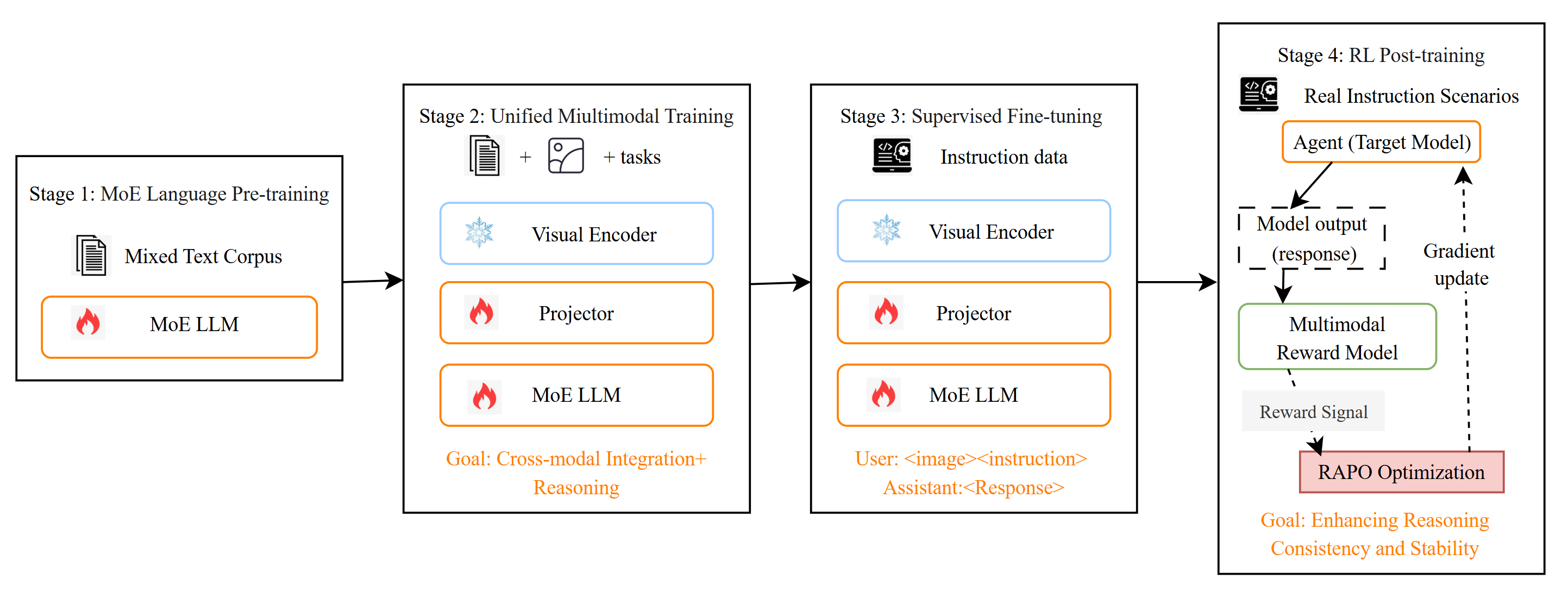}
  \caption{Structure of multi-stage collaborative multimodal training framework.}
  \label{fig:pipeline}
\end{figure}

The study constructs a unified, multi-stage collaborative multimodal training framework, designed to systematically achieve the alignment of visual and linguistic semantic spaces, the integration of cross-modal knowledge, and the enhancement of complex multimodal reasoning abilities in a stepwise manner. The overall process comprises four stages (Figure \ref{fig:pipeline}). 
First, the Yuan3.0 language model is pre-trained from scratch under a unified large-scale pre-training pipeline on about 3 trillion tokens. 
Second, the MoE language backbone and projector are unfrozen to perform unified multimodal training on 256 million image-text pairs.
The third stage involves supervised fine-tuning based on high-quality instruction data, further improving the model’s multimodal instruction-following and reasoning capabilities. 
Finally, large-scale reinforcement learning is conducted, continuously optimizing the model’s reasoning consistency and behavior quality in real interaction scenarios. The RL training adopted a hybrid training mode integrating thinking and non-thinking capabilities.

\section{Experiments}

\subsection{Long-Text Benchmarks}

\begin{figure}[t]
\centering
  \includegraphics[width=\linewidth]{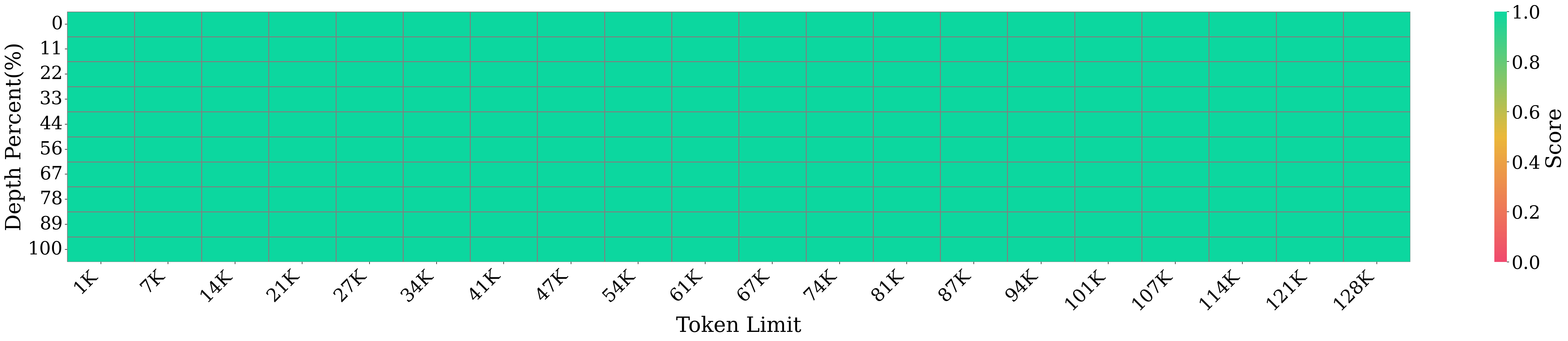}
  \caption{Evaluation results by NIAH test.}
  \label{fig:pressuretest}
\end{figure}

Yuan3.0 supports an extended context window of 128K tokens, enabling it to seamlessly process and reason over lengthy, complex documents. To assess the model’s ability on processing and understanding long textual data, we evaluate Yuan 3.0 Flash model using the Needle-in-a-Haystack (NIAH) benchmark \citep{xiao2024duoattention}. The experiments are conducted across a range of context lengths and needle positions, with response accuracy serving as the evaluation metric. The results are presented in Figure \ref{fig:pressuretest}. It can be concluded that our model achieves a perfect score within the 128K context scope. This indicates that the model can stably and accurately retrieve critical target information from ultra-long text corpora, which provides a reliable technical guarantee for its deployment in enterprise-level long-text application scenarios (e.g., long report analysis, multi-chapter knowledge reasoning).

\subsection{Enterprise scenarios benchmarks}

We further evaluate Yuan3.0 Flash on a suite of complex, enterprise-oriented tasks, covering multimodal retrieval, text retrieval, complex table understanding, text summarization, and tool invocation, which together reflect representative real-world application requirements.

For multimodal retrieval, we adopt the Docmatix benchmark \citep{laurenccon2024building} for multimodal RAG evaluation, which assesses a model’s ability to retrieve, associate, and accurately answer questions across multiple modalities (text, tables, and images) within multi-page, complex documents. On this benchmark, Yuan3.0 Flash achieves the highest performance among all of the models such as Claude 3.5, GPT-4o, and GPT-5.1 as shown is Table \ref{tab:Docmatix}. This result demonstrates strong multimodal reasoning and retrieval capabilities in document-level scenarios.

\begin{table}
  \centering
  \caption{Comparison among Yuan3.0 Flash and other representative models on Docmatix}
  \begin{tabular}{lc}
    \toprule
    \textbf{Model} & \textbf{Acc(\%)} \\
    \midrule
    Qwen2.5-VL-72B-Instruct     & 59.75           \\
    InternVL3-78B     & 42.99           \\
    Claude3.5-Sonnet     & 42.55          \\
    OpenAI GPT-4o     & 56.79           \\
    OpenAI GPT-o3      & 45.57            \\
   OpenAI GPT-5.1     & 48.52           \\
   Yuan3.0 Flash    & 65.07           \\
   \bottomrule
  \end{tabular}
  \label{tab:Docmatix}
\end{table}

For text retrieval, we evaluate on ChatRAG \citep{liu2024chatqa}, a widely recognized industrial benchmark comprising ten tasks. These tasks include long-context retrieval such as D2D, QuAC, QReCC, short-context and structured retrieval including CoQA, DoQA, CFQA, SQA, HDial, as well as Wikipedia-based retrieval tasks including TCQA, INSCIT. Across the full ChatRAG task set, Yuan3.0 Flash attains an average accuracy of 64.47, surpassing DeepSeek-V3, DeepSeek-R1, GPT-4o, and GPT-o3, and achieves leading performance on 9 out of the 10 tasks, indicating robust retrieval effectiveness across diverse textual contexts as shown in Table \ref{tab:chatrag}.

\begin{table}
  \centering
  \caption{Comparison among Yuan3.0 Flash and other representative models on ChatRAG (Acc. \%)}
  \resizebox{\textwidth}{!}{
    \begin{tabular}{lccccccccccc}
        \toprule
        \textbf{Models}  &\textbf{Avg All} &\textbf{D2D}  & \textbf{QuAC}  & \textbf{QReCC}  & \textbf{CoQA} & \textbf{DoQA} & \textbf{CFQA} & \textbf{SQA}  & \textbf{TCQA}  & \textbf{HDial}  & \textbf{INSCIT}\\
        \midrule
        DeepSeek-V3    &50.47	&31.59	&28.86	&49.31	&76.98	&26.11	&83.49	&82.13	&46.69	&47.43	&32.08 \\
        OpenAI GPT-4o   &50.54	&32.76	&26.56	&49.30	&76.11	&28.78	&81.85	&81.14	&49.75	&41.29	&26.69 \\                     
        OpenAI GPT-o3	&44.06	&23.05	&20.82	&40.42	&69.42	&18.56	&67.75	&86.71	&45.85	&41.29	&26.69 \\          
        DeepSeek-R1	&39.42	&21.46	&22.23	&42.41	&62.53	&24.68	&81.48	&82.06	&30.74	&37.97	&28.68     \\                      
        OpenAI GPT-5.1	&46.10	&28.24	&23.16	&45.43	&68.84	&20.88	&73.05	&81.32	&44.70	&45.39	&29.95     \\     
        Yuan3.0 Flash	&\textbf{64.47}	&\textbf{49.82}	&\textbf{53.79}	&\textbf{57.08}	&\textbf{90.93}	&\textbf{59.99}	&74.40	&\textbf{87.52}	&\textbf{66.31}	&\textbf{68.45}	&\textbf{36.40}  \\             
        \bottomrule
    \end{tabular}
}
  \label{tab:chatrag}
\end{table}

Multimodal table understanding is a critical capability in enterprise office scenario. We evaluate this ability using MMTab \citep{titiya2025mmtbench}, which consists of 15 authoritative benchmarks spanning multiple task categories. Firstly, question-answering tasks include the following benchmarks. TABMWP contains multiple table-based math word problems. WTQ and HiTab are the tasks of complex querying and reasoning over Wikipedia tables. TAT-QA is a benchmark for joint reasoning over financial tables and text. FeTaQA is a benchmarks for generating free-form answer from tables. Secondly, Fact-checking task includes TabFact, InfoTabs, HiTab T2T, Rotowire, and WikiBIO. Then, TSD, TCE, TCL, MCD, and RCE focus on long-context table-centric information processing. As shown in Table \ref{tab:MMtab}, Yuan3.0 Flash achieves a leading average accuracy of 58.29, exceeding GPT-5.1, and outperforms competing models on 7 of the 15 tasks, demonstrating comprehensive and well-balanced table reasoning and generation capabilities.

\begin{table}[!htbp]
\centering
\small 
\caption{Comparison among Yuan3.0 Flash and other representative models on MMTab} 
\label{tab:MMtab} 
\begin{tabular}{llcccc}
\toprule 
 &\textbf{Tasks} & \textbf{GLM-4.5V} & \textbf{OpenAI GPT-4V} & \textbf{OpenAI GPT-5.1} & \textbf{Yuan3.0 Flash} \\
\midrule
\textbf{Avg.}   &             & 52.00  & 29.90  & 55.15  & \textbf{58.29}  \\
\midrule
\multirow{5}*{\textbf{QA}}   &TABMWP (Acc.\%)     & 88.21  & 60.50  & 64.95  & \textbf{95.09}  \\
                    &WTQ (Acc.\%)        & 77.42  & 48.00  & 60.77  & 68.23  \\
                    &HiTab (Acc.\%)      & 51.52  & 27.50  & 77.77  & 69.80  \\
                    &TAT QA (Acc.\%)     & 62.69  & 32.50  & 61.37  & \textbf{69.17}  \\
                    &FeTaQA (BLEU)       & 5.25   & 11.04  & 8.70  & \textbf{28.42}  \\
\midrule
\multirow{5}*{\textbf{Fact Checking}} &TabFact (Acc.\%)    & 89.44  & 45.50  & 52.81  & 87.32  \\
                            &InfoTabs (Acc.\%)   & 79.48  & 65.60  & 64.30  & \textbf{83.50}  \\
                            &HiTab T2T (BLEU)    & 5.17   & 2.98   & 44.16  & 13.30  \\
                            &Rotowire (BLEU)     & 4.48   & 4.23   & 17.81  & 14.74  \\
                            &WikiBIO (BLEU)      & 2.69   & 1.94   & 11.95  & \textbf{17.26}  \\
\midrule
\multirow{2}*{\textbf{TSD}} &Row (Acc.\%)        & 47.40  & 19.00  & 96.60  & 46.60  \\
                            &Col (Acc.\%)        & 89.70  & 38.00  & 62.10  & 82.80  \\
\midrule
\textbf{TCE} (Acc.\%)  & & 52.74 & 14.36  & 86.43  & 56.77  \\
\textbf{TCL} (Acc.\%)  & &50.84 & 27.91  & 44.66  & \textbf{56.98}  \\
\textbf{MCD} (F1\%)  & &43.47  & 3.50   & 72.46  & 65.20  \\
\midrule
\multirow{2}*{\textbf{RCE}}  &Row (F1\%)  & 50.77  & 48.52  & 53.58  & \textbf{62.07}       \\
                             &Col (Acc.\%) & 82.79  &57.14  &57.20 &73.67\\
\bottomrule 
\end{tabular}
\end{table}

For text summarization, which is essential for compressing user history in agent-based applications, we evaluate on SummEval \citep{fabbri2021summeval}, covering lexical overlap, semantic similarity, and factual consistency. Yuan3.0 Flash achieves an average score of 59.31 as shown in Table \ref{tab:summeval}, outperforming Gemini (45.35) and GPT-5.1 (49.44), highlighting its strength in producing concise, semantically faithful, and factually consistent summaries.

\begin{table*}[t]
  \centering
  \caption{Comparison among Yuan3.0 and other representative models on SummEval.}
  \resizebox{0.8\textwidth}{!}{
  \begin{tabular}{l c c c c c}
    \toprule
    \textbf{Models} & \textbf{Avg. All} & \multicolumn{2}{c}{\textbf{Word Overlap}} & \textbf{Semantic} & \textbf{Factual} \\
    & \textbf{(W=100\%)} & \textbf{ROUGE-1} & \textbf{ROUGE-2} & \textbf{Similarity} & \textbf{Consistency} \\
    & & (F1\%) & (F1\%) & \textbf{BERTScore (F1\%)} & \textbf{SummaC (Acc. \%)} \\
    \midrule
    DeepSeek-V3 & 59.28 & 25.50 & 9.20 & 86.30 & 68.20 \\
    DeepSeek-V3.2 & 51.36 & 33.30 & 11.92 & 85.61 & 41.76 \\
    Gemini-2.0-Flash & 45.35 & 24.80 & 8.70 & 85.70 & 29.50 \\
    Claude-3.5-Sonnet & 45.43 & 24.10 & 8.30 & 85.20 & 30.70 \\
    OpenAI GPT-4o & 46.53 & 25.00 & 8.90 & 85.90 & 32.50 \\
    OpenAI GPT-5.1 & 49.44 & 27.48 & 10.16 & 84.63 & 40.50 \\
    Yuan3.0 Flash & \textbf{59.31} & \textbf{51.32} & \textbf{28.32} & \textbf{89.99} & 45.34 \\
    \bottomrule
  \end{tabular}
  }
  \label{tab:summeval}
\end{table*}

Finally, tool invocation ability is assessed using BFCL V3 \citep{patilberkeley}, which evaluates real-world function-calling competence across five tasks as shown in Table \ref{tab:bcfl}. Firstly,  Non-Live AST evaluates the ability of static function selection and argument extraction. Secondly, Live AST is a benchmark evaluating the ability of dynamic execution with real-time feedback. Thirdly, Multi-turn demostrates the ability of context maintenance and multi-tool coordination. Last, Relevance Detection evalutes deciding whether tool invocation is required), and Irrelevance Detection (rejecting invalid or unnecessary calls). Yuan3.0 Flash demonstrates consistently strong performance across all categories (57.97 on average) with no evident weaknesses, underscoring its reliability and maturity in tool-augmented reasoning and execution.

\begin{table*}[htbp]
  \centering
  \caption{Comparison among Yuan3.0 Flash and other representative models on BFCL.}
  \resizebox{\textwidth}{!}{
  \begin{tabular}{l c c c c c c}
    \toprule
    \textbf{Model} & \textbf{Avg. All} & \textbf{Non-Live AST} & \textbf{Live AST} & \textbf{Multi Turn} & \textbf{Relevance Detection} & \textbf{Irrelevance Detection} \\
    \midrule
    Qwen3-235B-A22B & 67.94 & 87.90 & 77.03 & 40.12 & 83.32 & 76.32 \\
    Claude-3.7-Sonnet & 58.58 & 41.29 & 78.41 & 48.38 & 72.22 & 81.40 \\
    OpenAI GPT-4.1-nano & 52.98 & 76.65 & 64.33 & 19.88 & 94.44 & 59.39 \\
    OpenAI o3-mini & 51.26 & 42.12 & 77.30 & 26.12 & 77.78 & 80.67 \\
    Llama-4-Scout-17B-16E-Instruct & 45.41 & 83.48 & 57.97 & 1.88 & 100.00 & 39.66 \\
    Yuan3.0 Flash & 57.97 & 69.06 & 66.37 & 36.88 & 94.44 & 77.09 \\
    \bottomrule
  \end{tabular}
  }
  \label{tab:bcfl}
\end{table*}

Overall, these results indicate that Yuan3.0 Flash delivers competitive and often leading performance across a broad spectrum of enterprise-level complex tasks, with particular strengths in multimodal retrieval, table understanding, summarization quality, and robust tool-calling behavior.

\subsection{Multimodal Benchmarks}
To comprehensively evaluate the capabilities of Yuan3.0 Flash model across diverse visual-language understanding scenarios, we evaluate on the following benchmarks.
AI2D \citep{kembhavi2016diagram} is a benchmark for reasoning over science diagrams, requiring models to parse visual elements and their relationships to answer questions. ChartQA \citep{masry2022chartqa} evaluates chart comprehension by requiring models to extract and reason about data from various chart types (e.g., bar, line, pie) to answer factual and reasoning questions. DocVQA \citep{mathew2021docvqa} tests the understanding of scanned documents, requiring models to answer questions by interpreting textual content, layout, and visual cues within document images. MathVista \citep{lu2023mathvista} is a benchmark designed to evaluate mathematical reasoning in visual contexts, combining problems from various existing datasets that require jointly understanding figures, plots, and diagrams to perform mathematical calculation and reasoning.

\begin{table}[htbp]
\centering
\caption{Comparison among Yuan3.0 Flash and other models on visual benchmarks}
    \begin{subtable}{0.9\textwidth}
      \centering
      \caption{Comparison of performance in Non-Thinking settings}
      \begin{tabular}{lccc}
        \toprule
           &\makecell[c]{Yuan3.0 Flash \\ Non-Thinking Mode}   &\makecell[c]{Qwen3-VL-32B- \\ Instruct}  &\makecell[c]{Qwen2.5-VL- \\ 72B}  \\
        \midrule
       ChartQA	 &87.9	&88.5	&89.5     \\
        DocVQA	&90.1	&96.9	&96.4      \\
        AI2D	&85.7	&89.5	&88.7    \\
        MathVista	&72.8	&83.8	&74.8 \\
       \bottomrule
      \end{tabular}
      \label{tab:visualfast}
    \end{subtable}
    \hfill
    \begin{subtable}{0.9\textwidth}
      \centering
      \caption{Comparison of performance in Thinking settings}
      \begin{tabular}{lccc}
        \toprule
           &\makecell[c]{Yuan3.0 Flash \\ Thinking Mode}   &\makecell[c]{Qwen3-VL-32B- \\ Thinking}  &\makecell[c]{Qwen3-VL-235B- \\ A22B-Thinking}  \\
        \midrule
       ChartQA	 &90.1	&89.0	&90.3     \\
        DocVQA	&90.2	&96.1	&96.5      \\
        AI2D	&86.9	&88.9	&89.2   \\
        MathVista	&74.1	&85.9	&85.8 \\
       \bottomrule
      \end{tabular}
      \label{tab:visualthink}
    \end{subtable}
    \hfill
    \begin{subtable}{0.9\textwidth}
      \centering
      \caption{Comparison of average number of tokens in Thinking settings}
      \begin{tabular}{lccc}
        \toprule
           &\makecell[c]{Yuan3.0 Flash \\ Thinking Mode}   &\makecell[c]{Qwen3-VL-32B- \\ Thinking}  &\makecell[c]{Qwen3-VL-235B- \\ A22B-Thinking}  \\
        \midrule
       ChartQA	 &341	&602	&802     \\
        DocVQA	&180	&217	&375      \\
        AI2D	&427	&1,234	&1,387   \\
        MathVista	&885	 &1,761	&1,665 \\
       \bottomrule
      \end{tabular}
      \label{tab:visualthinktoken}
    \end{subtable}
\end{table}

As shown in Table \ref{tab:visualfast}, we evaluate models in non-thinking mode across four tasks. For ChartQA and DocVQA, YUAN3.0 Flash performs close to Qwen3-VL-32B and Qwen2.5-VL-72B, demonstrating solid abilities on OCR and document scenarios. For AI2D, the performance of Yuan3.0 Flash is silightly behind other models but has comparable scores, reflecting the ability of the model in handling structured and scientific reasoning tasks. In MathVista task, Yuan3.0 Flash scored 72.5, which is close to Qwen2.5-VL-72B but behind Qwen3-VL-32B. The comparisons prove the capabilities in mathematical computation and logical reasoning. As for thinking mode, the performance of Yuan3.0 Flash Thinking Mode showed gains in some tasks compared to its non-thinking mode. Comparing with other models, Yuan3.0-40B has higher performance than Qwen3-VL-32B-thinking and is close to Qwen3-VL-235B-A22B on ChartQA as shown in Table \ref{tab:visualthink}. For AI2D task, Yuan3.0 Flash achieves a slight increase in non-thinking mode and comes close to Qwen3-VL-235B-A22B and Qwen3-VL-32B. In the MathVista task, Yuan3.0 Flash obtained a score higher than the non-thinking score, but has a lower score than Qwen3-VL-235B-A22B and Qwen3-VL-32B. To further evaluate and compare the computational efficiency of the reasoning process and its trade-off with accuracy, we collected statistics on the average number of output tokens generated during inference for our model and other baseline models. As shown in Table \ref{tab:visualthinktoken}, Yuan3.0 Flash Thinking Mode significantly reduces computational cost compared to the two Qwen3 models. Specifically, its token count represents a reduction to between 1/3 and 1/2 of that of Qwen3-VL-235B-A22B, effectively mitigating overthinking issue and reducing unnecessary resource consumption in the reasoning process.

In summary, the minimal performance gap between its two modes of YUAN3.0 Flash highlights the practical efficiency of the lightweight non-thinking design. In addition, Yuan3.0 Flash Thinking Mode maintains accuracy comparable to Qwen3 series models while significantly reduces the token consumption.

\subsection{Evaluation on General Reasoning Benchmarks}
To evaluate mathematical, coding, etc., reasoning capability, we conduct benchmarks on MATH-500 \citep{lightman2023let}, AIME 2024 \citep{codeforcesamerican}, LiveCodeBench \citep{jain2024livecodebench}, HumanEval \citep{chen2021evaluating}, MMLU, MMLU-Pro\citep{wang2024mmlu}, and GPQA-Diamond \citep{rein2024gpqa}. We averaged 64 samples per question for AIME, while 8 samples per question for others.

\begin{table}
  \centering
  \caption{Accuracy comparison of Yuan3.0 Flash and DeepSeek-V3-0324 under non-thinking mode}
  \resizebox{0.5\textwidth}{!}{
  \begin{tabular}{lcc}
    \toprule
      Benchmarks &\makecell[c]{DeepSeek-V3-0324  \\ 671B}  &\makecell[c]{Yuan3.0 Flash \\40B} \\
    \midrule
   AIME 2024	&59.4	&32.6     \\
    MATH-500	&94.0	 &88.7      \\
   LiveCodeBench	&35.3	&22.5   \\
    HumanEval	&92.9	&86.8 \\
    GPQA-Diamond	&68.4	&37.4 \\
    MMLU	&83.4	&82.9  \\
    MMLU pro	&81.2	&64.2 \\
   \bottomrule
  \end{tabular}
  }

  \label{tab:textnonthink}
\end{table}


\begin{table}[htbp]
\centering
\small
\setlength{\tabcolsep}{3pt}
\caption{Accuracy comparison of Yuan3.0 Flash and DeepSeek-R1-0528 under thinking mode.}
\begin{tabular}{@{}l*{14}{c}@{}}
\toprule 
\multirow{2}{*}{Model} & \multicolumn{2}{c}{AIME 2024} & \multicolumn{2}{c}{MATH-500} & \multicolumn{2}{c}{LiveCodeBench} & \multicolumn{2}{c}{HumanEval} & \multicolumn{2}{c}{GPQA-Diamond} & \multicolumn{2}{c}{MMLU} & \multicolumn{2}{c}{MMLU pro} \\
\cline{2-15}
                        & Acc & Tokens & Acc & Tokens & Acc & Tokens & Acc & Tokens & Acc & Tokens & Acc & Tokens & Acc & Tokens \\
\midrule
DeepSeek-R1-0528        & 91.4 & 17,164 & 97.4 & 5,541 & 73.3 & 17,896 & 96.9 & 3,135 & 81.0 & 10,081 & 88.7 & 1,616 & 85.0 & 4,501  \\
Yuan3.0 Flash 40B       & 47.6 & 6,086  & 91.2 & 1,431 & 29.8 & 8,157  & 95.6 &  2,313 & 47.4 & 3,761 & 83.7 & 806 & 68.1 & 1,688 \\
\bottomrule
\end{tabular}
  \label{tab:textthink}
\end{table}

The Table \ref{tab:textnonthink} presents comparison of the Yuan3.0 Flash model with Deepseek-V3-0324 under non-thinking mode. Yuan3.0 Flash (40B) demonstrates performance gaps relative to DeepSeek-V3-0324 (671B) on AIME2024, GPQA-Diamond, and MMLU pro.
For MMLU, the gap shrinks to just 0.5 percentage points (82.9\% vs. 83.4\%). For MATH-500, HumanEval, LiveCodeBench, the accuracy of Yuan3.0 Flash is close to DeepSeek-V3-0324. These findings establish Yuan3.0 Flash as a competitive alternative to larger models in core reasoning tasks, delivering comparable accuracy with a fraction of the parameters.

In the thinking mode, we analyze the performance of Yuan3.0 Flash compared with DeepSeek-R1-0528 across a wide range of benchmarks as shown in Table \ref{tab:textthink}. For MATH-500, HumanEval, MMLU, the accuracy of Yuan3.0 Flash is close to DeepSeek-R1-0528, while consuming up to nearly 1/4 tokens. For AIME2024, GPQA-Diamond, and MMLU pro, the accuracy lags behind DeepSeek-R1-0528, while the tokens of Yuan3.0 Flash is up to around 1/3 tokens. In conclusion, the results establish that Yuan3.0 Flash is robustly capable of complex reasoning with superior token efficiency.

\section{Conclusion}
In this paper, we introduce Yuan3.0 Flash, a 40B MoE multimodal language model with hybrid of thinking and non-thinking mode, is built to enhance performance on enterprise tasks while maintaining competitive capabilities on general purpose tasks. We propose RAPO algorithm that effectively mitigates the overthinking issue of LRMs, significantly improves training efficiency and model accuracy. Yuan3.0 Flash demonstrates superior performance in enterprise scenarios such as RAG, complex table processing, etc. Moreover, it also demonstrates strong reasoning capabilities in mathematics, science, etc., attaining accuracy comparable to frontier models while requiring only approximately 1/4 to 1/2 of the average tokens.

\section{Contribution}
Shawn Wu, Sean Wang, Louie Li, Darcy Chen, Allen Wang, Jiangang Luo, Xudong Zhao, Joseph Shen, Gawain Ma, Jasper Jia, Marcus Mao, Claire Wang, Hunter He, Carol Wang, Zera Zhang, Jason Wang, Chonly Shen, Leo Zhang, Logan Chen, Qasim Meng, James Gong, Danied Zhao, Penn Zheng, Owen Zhu, Tong Yu

\bibliographystyle{unsrt} 
\bibliography{references}  

\appendix

\section{Adaptive Image Segmentation for High-Resolution Visual Inputs}
\label{app:seg}

To mitigate geometric distortion and fine-grained detail loss when processing high-resolution images in MLLMs, we adopt an \textbf{Adaptive Image Segmentation Strategy} formulated as a dynamic programming optimization problem. Given an input image with resolution $(W_I,H_I)$ and the fixed input resolution of the visual encoder $(W_v,H_v)$, the algorithm automatically determines the optimal number of image slices and their grid configuration. The objective is to maximally preserve the original geometric structure while maintaining computational efficiency.

To accommodate diverse image aspect ratios, we predefine a set of allowable slice counts,
\begin{equation}
    N_{all} = \left \{ 1,2,… ,9 \right \} 
\end{equation}
For each candidate slice number $N_{i} \in  N_{all} $, all feasible grid configurations $(m,n)$ satisfying $m\times n=N_{i}$ are enumerated. This yields a collection of candidate segmentation schemes:
\begin{equation}
    C_{N}=\left \{ m\times n=N_{i}, N_{i}\in  N_{all} \right \}  
\end{equation}
where $\textit{m}$ denotes the number of columns (horizontal partitions) and $\textit{n}$ denotes the number of rows (vertical partitions).

To ensure that the segmented patches preserve the geometric proportions of the original image and to minimize distortions introduced by non-uniform resizing, we define a shape consistency score $S\left ( \cdot  \right ) $ that measures the discrepancy between the aspect ratio of the original image and that of the segmentation grid:
\begin{equation}
S\left (W_{l},H_{l},m,n\right ) =\left | \frac{W_{l}}{H_{l}}-\frac{m}{n} \right | 
\end{equation}
A smaller $S$ indicates a closer match between the grid layout and the original image aspect ratio, and thus lower geometric distortion.

However, optimizing shape consistency alone may result in excessive segmentation, particularly for small images, leading to unnecessary up-sampling and increased computational overhead. To address this issue, we introduce a regularization threshold $\tau$ to constrain the expansion ratio. Specifically, we define the expansion ratio difference:
\begin{equation}
    d\left ( m,n \right ) =\left | \frac{\left ( W_{v}\times n + H_{v}\times m  \right )-\left (W_{l}+H_{l}  \right )  }{W_{l}+H_{l}}  \right | 
\end{equation}
Only candidate schemes satisfying $ d\left ( m,n \right )< \tau $ are retained, forming a filtered candidate set $C_{N}^{'}$.

If $C_{N}^{'} \neq \varnothing$, the input image is considered too small to benefit from segmentation, and a default $(1,1)$ configuration is applied, i.e., no segmentation is performed. Otherwise, the optimal segmentation scheme $\left ( m^{*}, n^{*}  \right )$ is selected by minimizing the shape consistency score:
\begin{equation}
    \left ( m^{*}, n^{*}  \right ) = argmin_{\left (m,n  \right )\in C_{N}^{'} } S\left ( W_{l}, H_{l},m,n \right )
\end{equation}
In cases where multiple candidate schemes yield identical $S$ values, a tie-breaking rule is applied based on the effective resolution after segmentation. Specifically, if a scheme results in a total segmented area less than 50\% of the original image area, i.e.,
\begin{equation}
    0.5\times \left ( W_{v}\times H_{v} \times n \times m \right ) < W_{l}\times H_{l}
\end{equation}
it is preferred, as this avoids ineffective segmentation that primarily introduces redundant up-sampling.

Once the optimal segmentation parameters $\left ( m^{*}, n^{*}  \right )$ are determined, the following preprocessing steps are performed:
\begin{enumerate}
    \item \textbf{Local Slices:} The original image is partitioned into an $\left ( m^{*}, n^{*}  \right )$ grid, and each local patch is resized to the visual encoder’s input resolution $(W_v,H_v)$.
    \item \textbf{Global Thumbnail:} The original image is directly resized to $(W_v,H_v)$ to provide a global contextual view.
    \item \textbf{Feature Concatenation:} All local patches and the global thumbnail are concatenated to form the final visual input sequence for the multimodal model.
\end{enumerate}

This adaptive image segmentation strategy enables efficient utilization of high-resolution visual information while preserving both global structure and local details, thereby improving visual grounding in MLLMs.

\section{Adaptive Dynamic Sampling}
\label{app:ADS}

\begin{algorithm}[H]
\caption{Adaptive Dynamic Sampling}
\label{alg:ads}
\begin{minipage}{0.8\textwidth}
\KwIn{\\
    \hspace*{2em}Train batch $\mathcal{B}$; Global batch size $gbs$ (eg. 512); Mini batch size $mbs$ (eg. 64) \\
}
\KwOut{\\
    \hspace*{2em}resampled train batch $B_r$}
\BlankLine

Initialize prompts buffer $Q$ for filtered prompts

Initialize pass rates buffer $P_a$ for pass rates of $Q$
\BlankLine

\textbf{Step 1: Filter train batch $\mathcal{B}$}\\
\For{$q_j$ in $\mathcal{B}$ }{
    \If{all outputs of $q_j$ are correct or incorrect}{
        filter $q_j$
    }
    \Else{
        $q_j \rightarrow Q$\\
        $p_j=\frac{1}{G}\sum_{i=1}^G\mathbb{I}(a_i=1)$ \Comment{Compute pass rate of $q_j$}\\
        $p_j \rightarrow P_a$
    }
}
\BlankLine

\textbf{Step 2: Sort $Q$ by pass rate}\\
Sort $Q$ in descending order of pass rate
\BlankLine

\textbf{Step 3: Compute the target size for completion}\\
$k = \lfloor \frac{|Q|+mbs-1}{mbs} \rfloor$\Comment{Compute the smallest multiples of $mbs$}\\
$T = k \times mbs$\Comment{Compute target batch size}
\BlankLine

\textbf{Step 4: Complete train batch by resampling}\\
select the top $N$ prompts $Q^*$ in $Q$\\
$combine(Q^*,Q)\rightarrow \mathcal{B}_r$\\
\Return{$\mathcal{B}_r$}
\end{minipage}
\end{algorithm}

\end{document}